\documentclass[final]{cvpr}

\usepackage{times}
\usepackage{epsfig}
\usepackage{graphicx}
\usepackage{amsmath}
\usepackage{amssymb}
\usepackage{bbm}
\usepackage{enumitem}

\usepackage[pagebackref=true,breaklinks=true,colorlinks,bookmarks=false]{hyperref}

\def\cvprPaperID{5808} 
\def\confYear{CVPR 2021}

\begin{document} 


\title{DARCNN: Domain Adaptive Region-based Convolutional Neural Network for Unsupervised Instance Segmentation in Biomedical Images}

\author{Joy Hsu\\
Stanford University\\
{\tt\small joycj@stanford.edu}
\and
Wah Chiu\\
Stanford University\\
{\tt\small wahc@stanford.edu}
\and
Serena Yeung\\
Stanford University\\
{\tt\small syyeung@stanford.edu}
}

\maketitle

\begin{abstract}
In the biomedical domain, there is an abundance of dense, complex data where objects of interest may be challenging to detect or constrained by limits of human knowledge. Labelled domain specific datasets for supervised tasks are often expensive to obtain, and furthermore discovery of novel distinct objects may be desirable for unbiased scientific discovery. Therefore, we propose leveraging the wealth of annotations in benchmark computer vision datasets to conduct unsupervised instance segmentation for diverse biomedical datasets. The key obstacle is thus overcoming the large domain shift from common to biomedical images. We propose a Domain Adaptive Region-based Convolutional Neural Network (DARCNN), that adapts knowledge of object definition from COCO, a large labelled vision dataset, to multiple biomedical datasets. We introduce a domain separation module, a self-supervised representation consistency loss, and an augmented pseudo-labelling stage within DARCNN to effectively perform domain adaptation across such large domain shifts. We showcase DARCNN's performance for unsupervised instance segmentation on numerous biomedical datasets.
\end{abstract}

\begin{figure}[htb]
\begin{center}
\includegraphics[width=1.0\linewidth]{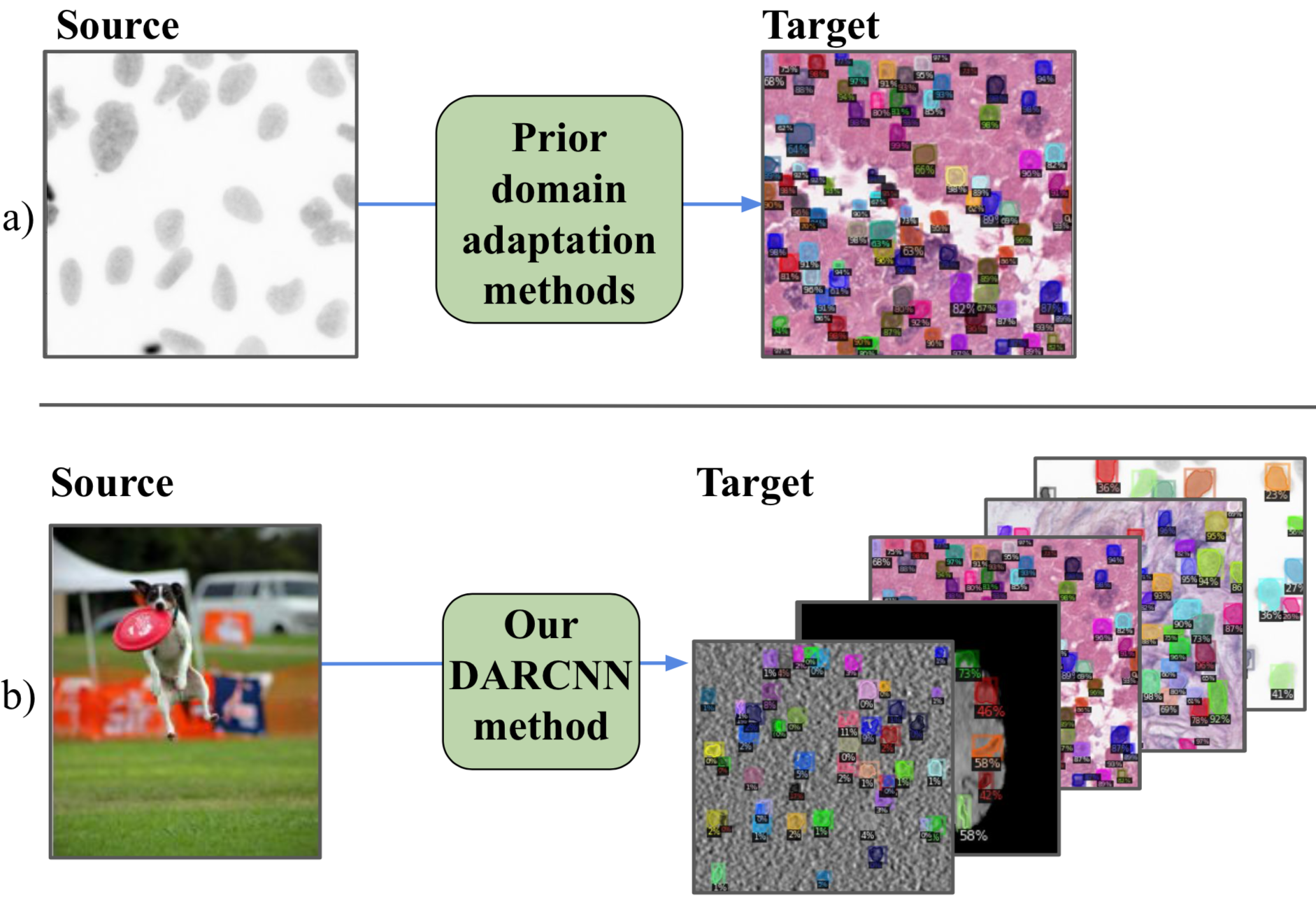}
\end{center}
   \caption{a) Prior domain adaptation methods for biomedical images tackle small domain shifts by using similar labelled biomedical datasets as sources to adapt to specific target datasets. b) DARCNN uses a common benchmark dataset as source and can adapt to a wide range of biomedical images.}
\label{fig:pull}
\vspace{-1em}
\end{figure}

\vspace{-1em}
\section{Introduction}
State-of-the-art machine learning methods have accomplished a wide variety of impressive tasks including instance segmentation, yet much of their progress in the real world is limited to supervised methods with large, labelled datasets. In areas such as the biomedical domain, this is particularly problematic, as the prerequisite labels that accompany the complex data are often time consuming to obtain. In addition, we may also be constrained by human knowledge --- biomedical data often contains unknown objects that scientists have yet to uncover, and therefore cannot accurately annotate. 

Thus, there exists a need for methods that can produce instance segmentation for unlabelled datasets. We tackle this problem through solving the unsupervised domain adaptation task, in which we use a source dataset with instance segmentation annotations to transfer knowledge and perform instance segmentation on target datasets.

Our choice of source dataset is motivated by the abundance of benchmark datasets in the vision field depicting common objects. We explore leveraging the large amount of labelled vision data in Common Objects in Context (COCO) \cite{lin2014microsoft} to achieve instance segmentation in diverse, natural biomedical images where annotations are difficult to obtain. Our main contributions include overcoming the large domain shift between natural images and biomedical images, and introducing a method for unsupervised instance segmentation on a wide range of biomedical datasets.

Past work tackling this problem in the biomedical field have depended on the availability of similar labelled biomedical datasets for the unsupervised instance segmentation task (see Figure~\ref{fig:pull}), but it is not always feasible to find and annotate similar images. For these prior domain adaptation methods that focus on small domain shifts, joint image-level and feature-level adaptation approaches and object-specific models have seen success \cite{chen2019synergistic, chen2018domain, hoffman2018cycada, kim2019diversify}. However, few works study unsupervised domain adaptation on large domain shifts such as from COCO to biomedical images, where such image-level adaptation fails. In addition, other past methods also design models specific to segmenting particular structures, which limits both application to other biomedical datasets as well as discovery \cite{hou2019robust, liu2020unsupervised}. 



Hence we propose Domain Adaptive Region-based Convolutional Neural Network (DARCNN), a two stage class agnostic unsupervised domain adaptation model for instance segmentation of all distinct objects, capturing the notion of objectness. DARCNN first tackles feature-level adaptation, then refines segmentation masks through image-level pseudo-labelling. Our method can be applied to datasets with consistent background (e.g. of homogeneous cell background in microscopy) instead of split backgrounds (e.g. of the sky and grass as commonly seen in COCO). DARCNN leverages the success of the two step Mask R-CNN framework \cite{he2017mask} and learns domain invariant and specific features for region proposal and segmentation mask prediction. The features are learned through a self-supervised background representation consistency loss based on predicted regions within an image.


In the second stage of DARCNN, pseudo-labelling on augmented input is introduced as a strong supervisory image-level signal. Through pseudo-labelling we are able to attain stable image-level segmentation after feature-level adaptation. We discover that our sequential two stage process is able to solve the domain adaptation task with large concept shift, shown on several biomedical datasets. In addition, we demonstrate that our method achieves strong performance on tasks of smaller domain shift as well.

Our key contributions are the following:
\begin{itemize}[noitemsep]
\item We introduce a domain separation module to learn domain invariant and domain specific features for the two step instance segmentation framework.
\item We propose a self-supervised representation consistency loss based on predicted regions within an image for feature adaptation.
\item We utilize pseudo-labelling with data augmentation within DARCNN for strong image-level supervision.
\item We demonstrate the effectiveness of our approach through quantitative experiments on adapting from COCO to five diverse biomedical datasets and a qualitative experiment for object discovery on a cryogenic electron tomography dataset.
\end{itemize}

\section{Related Work}
\subsection{Unsupervised Domain Adaptation}
Prior unsupervised domain adaptation approaches can be categorized into feature-level adaptation, image-level adaptation, or a combination of both. Feature-level adaptation includes minimizing distances between source and target features through extracting shared domain features \cite{ganin2015unsupervised, tzeng2017adversarial}, minimizing maximum mean discrepancy \cite{long2015learning}, or adversarial approaches such as \cite{sun2015return, wen2019exploiting}. 

Image-level adaptation, such as those that tackle pixel-to-pixel translation between source and target domains, are often evaluated on adaptations between similar domains with no concept shift, such as from Cityspaces \cite{cordts2016cityscapes} to GTA \cite{richter2016playing}. Common works include \cite{isola2017image, zhu2017unpaired}, which conducts image-to-image translation through generative adversarial networks. However, approaches such as as pixel-to-pixel translation are extremely limited by size of domain shift. 

Several key domain adaptive methods formulates adaptation across both image-level and feature-level, including \cite{hoffman2018cycada, kim2019diversify}, while others approach domain shift on the instance-level and image-level as well \cite{chen2018domain}. These methods conduct feature-level and image-level adaptation jointly or rely heavily on image-level adaptation, which has seen impressive results in adaptation with small domain shifts, but struggle with larger concept shifts. DARCNN conducts first feature-level adaptation then image-level refinement sequentially in a two stage process, overcoming limitations of prior works.

In addition, a line of prior work that tackles small domain shift across biomedical images has shown strong performance on specific datasets. \cite{chen2019synergistic} transforms appearance of MR and CT images through synergistic fusion of adaptations from both feature-level and image-level, and \cite{hou2019robust} generates synthesized nucleus segmentation masks with importance weighting. \cite{liu2020unsupervised} similarly introduces a nuclei inpainting mechanism for unsupervised nucleus segmentation through domain adaptation with re-weighting. However, these works tackle specific biomedical datasets, where methods can be crafted for detection of specific objects, and are also limited to smaller domain shifts, where an additional labelled biomedical dataset is needed for adaptation.

Our work overcomes the limitations of small domain shift and object-specific techniques for biomedical datasets through a two stage feature-level adaptation and image-level pseudo-labelling that segments all objects of interests. Previous works such as \cite{bousmalis2016domain} has shown success in domain separation networks in simpler classification settings; similarly, we introduce a domain adaptation module and integrate this with a self-supervised loss that learns feature discriminability for instance segmentation.

\subsection{Unsupervised Background-Foreground Segmentation}
Prior works on unsupervised background-foreground segmentation primarily use a combination of consistency constraints and domain-specific assumptions. For example, \cite{winn2005locus} focuses on consistency between generated image and outputs of edge detectors, \cite{rubinstein2013unsupervised} leverages salient pixels in the foreground and matching foregrounds between different images, and \cite{benny2019onegan} utilizes a multi-task formulation with need for clean background images.

DARCNN similarly takes a self-supervised approach by maintaining a background representation consistency constraint, leveraging proposed regions within each image. Through this objective, our approach is able to learn domain invariant and domain specific representation for segmentation.

\subsection{Pseudo-labels}
Pseudo-labelling has often been used as a technique for utilizing unlabelled data in semi-supervised training. Prior work have chosen maximum predicted probability labels \cite{lee2013pseudo}, uncertainty weighted class predictions \cite{shi2018transductive}, and used group-based label propagation \cite{iscen2019label}. Similarly, co-training methods use an ensemble of models to find labels through consistency regularization \cite{qiao2018deep}. Fewer methods have used this method in the unsupervised setting, though works such as \cite{choi2019pseudo} have used high density clusters as pseudo-labels, inferring high confidence without supervision.

After first stage feature-level adaptation, our unsupervised method uses pseudo-labels to gain stronger image-level supervision, where high confidence pseudo-labels comes from first stage DARCNN. In addition, to learn across invariances, we add data augmentation to the unlabelled target images such as in \cite{berthelot2019mixmatch}, allowing DARCNN to learn across different imaging conditions.

\section{Methods}
We propose methods for unsupervised instance segmentation through the task of domain adaptation with large concept shift. In this section, we describe our two stage DARCNN model. The initial stage of feature-level adaptation consists of a domain separation module and a self-supervised representation consistency loss, which can be found in Section~\ref{section:domain_separation_module} and Section~\ref{section:self_supervised}. The second stage of image-level pseudo-labelling can be found in Section~\ref{section:pseudo_labelling}.

\begin{figure*}[htb]
\begin{center}
\includegraphics[width=0.95\linewidth]{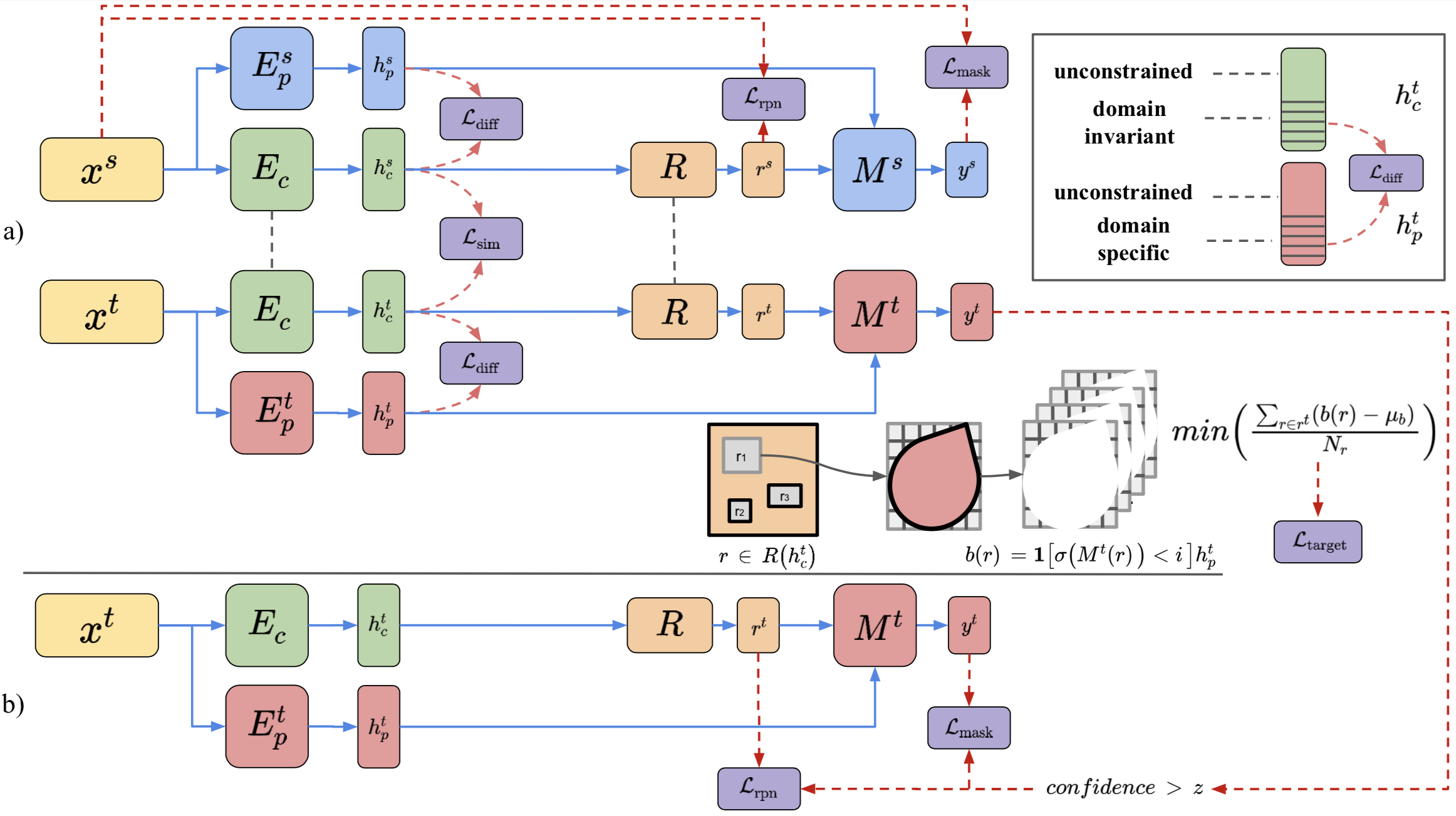}
\end{center}
   \caption{a) First stage DARCNN model with a domain separation module and self-supervised representation consistency loss. Let $s$ and $t$ represent source and target, and $h_c$ and $h_p$ represent common domain invariant features and private domain specific features respectively. $E_c$ is the shared encoder, $E_p^s$ and $E_p^t$ are domain specific encoders, $R$ is the shared region proposal network, and $M^s$ and $M^t$ are domain specific mask prediction heads. We let $b(r)$ be extracted background features for each region $r$. Top right corner showcases our soft orthogonality constraint on half of the domain specific features. b) Second stage DARCNN model with pseudo-labels of augmented input from first stage DARCNN, with annotations chosen over a confidence threshold, continuing training of DARCNN's target branch.}
\label{fig:systems}
\vspace{-1em}
\end{figure*}

We train each stage separately to tackle our problem of large domain shift. Image-level adaptation such as pixel-to-pixel translation does not work on such extreme concept shift, thus sequentially using image-level pseudo-labelling as a second stage allows for features to first learn to adapt between domains, before augmenting training with stronger pixel-level supervisory signal. DARCNN is pre-trained with source dataset weights, and jointly trains with a batch of source and target inputs. See Figure~\ref{fig:systems} for an overview of our model.

\subsection{Domain Separation Module}
\label{section:domain_separation_module}
The Mask R-CNN framework \cite{he2017mask} is a powerful instance segmentation model, and we leverage its two step framework for DARCNN as well as propose a domain separation module designed for our task. The region proposal network from the first step finds potential bounding boxes of interest given features learned through convolutional layers, while the mask prediction head from the second step refines these boxes and produces a mask for each instance. 

To tackle the problem of domain shift, we propose a domain separation module that learns domain invariant and domain specific features as input into the region proposal network and mask segmentation network. The domain invariant features encode objectness of the source and target domain in a joint representational subspace, while the domain specific features capture discriminability of each domain as well as contain additional unconstrained embedding space.



The losses of the DARCNN are: $L_{\textrm{sim}}$ to encourage domain invariant features, $L_{\textrm{diff}}$ to learn domain specific features, $L_{\textrm{source}}$ which includes the original Mask R-CNN losses for supervised training of the source dataset, and our proposed $L_{\textrm{target}}$ for segmentation through a self-supervised consistency loss. Weighting factors $\alpha$, $\beta$, and $\gamma$ are used to balance the loss. See Equation~\ref{eqn:all_losses} below.

\vspace{-1em}
\begin{align}\label{eqn:all_losses} 
    L_{\textrm{DARCNN}} = \alpha L_{\textrm{sim}} + \beta L_{\textrm{diff}} + \gamma L_{\textrm{target}} + L_{\textrm{source}}
\end{align}

\subsubsection{Domain Invariant Features}
Intuitively, region proposals should be based on high level definition of objectness in the input image shared between both domains. Hence we encourage source and target domain invariant features to move into a joint representational subspace. The similarity loss helps the unlabelled target features better encode the objectness learned from the labelled source features. We utilize the maximum mean discrepancy loss $L_{\textrm{sim}}$ from Equation~\ref{eqn:sim} below. The maximum mean discrepancy loss \cite{long2015learning, bousmalis2016domain} is a kernel-based distance function between pairs of samples; we can think of the loss as computing the difference in distribution $s$ and $t$ where source inputs are drawn from $s$ and target inputs are drawn from $t$. We let $\kappa$ be our kernel function, and $h_c^s$ and $h_c^t$ be our shared source and shared target features respectively.

\vspace{-1em}
\begin{align}\label{eqn:sim}
    L_{\textrm{sim}} & = \frac{1}{(N^{\textrm{s}})^2} \sum_{i, j = 0}^{N^{\textrm{s}}} \kappa (h_c^{s, i}, h_c^{s, j})  \\
    & - \frac{2}{N^{\textrm{s}} N^{\textrm{t}}} \sum_{i, j = 0}^{N^{\textrm{s}} N^{\textrm{t}}} \kappa (h_c^{s, i}, h_c^{t, j}) 
    + \frac{1}{(N^{\textrm{t}})^2} \sum_{i, j = 0}^{N^{\textrm{t}}} \kappa (h_c^{t, i}, h_c^{t, j}) \nonumber
\end{align}

In our implementation of maximum mean discrepancy, we use a Gaussian kernel $\kappa$. We downsample $h_c^s$ and $h_c^t$ with a dimension reduction convolutional layer with 1 filter of 1x1 kernel as an additional projection head to learn source and target domain invariant features.

However, as we focus on scenarios where domain shift between our source and target domains is large, the initial distance between $s$ and $t$ as calculated by the maximum mean discrepancy loss is also large. \cite{wang2020rethink} shows that minimizing the maximum mean discrepancy loss equates to maximizing the source and target intra-class distances respectively, but doing so also jointly minimizes their variance with some implicit weights, such that feature discriminability degrades. Therefore, if $L_{\textrm{sim}}$ too quickly overwhelms other losses in DARCNN that retain semantic features, the discriminability needed for instance segmentation is lost.

Hence we propose a maximum mean discrepancy loss that uses a warmup weighting scheduler. Our approach increases the weight $\alpha$ of $L_{\textrm{sim}}$ from $\alpha_0$ to $\alpha$ over $n$ epochs, where $\alpha_0$ is smaller when the domain shift is larger.



\subsubsection{Domain Specific Features}
We next consider input needed for mask predictions. Domain specific features captures feature discriminability for the target and source domains as well as granularity of the background representation for our self-supervised loss. Hence we use $L_{\textrm{diff}}$ to separate information that is unique to each domain as well as learn specificity, and define the loss through the soft subspace orthogonality constraint from \cite{bousmalis2016domain} between the domain invariant and domain specific features of both domains.

However, in order to let part of the mask feature representations learn semantically relevant embeddings that may potentially be domain invariant, we utilize the orthogonality difference loss only between parts of the domain invariant and domain specific features. We give DARCNN the freedom to learn features necessary for segmentation in an unconstrained embedding space, whether it be domain invariant or specific. In our implementation, we use half of the feature depth. See top right corner of Figure~\ref{fig:systems}. We let $H$ be matrices whose rows are half of hidden representations $h$ in depth, where $h_c^s$ and $h_p^s$ are the invariant and specific features of the source, and $h_c^t$ and $h_p^t$ of the target.

\vspace{-1em}
\begin{align}\label{eqn:diff}
L_{\textrm{diff}} = || H_c^{s \top} H_p^s ||^2_F + || H_c^{t \top} H_p^t||^2_F
\end{align}

To additionally give signal to the unlabelled target domain specific features, we propose a self-supervised representation consistency loss instead of the reconstruction loss used for classification in \cite{bousmalis2016domain}. The source domain specific features are supervised by the original Mask R-CNN bounding box and mask losses.

\subsection{Self-Supervised Representation Consistency}
\label{section:self_supervised}
Biomedical images commonly contain homogeneous backgrounds, therefore we leverage this assumption between region proposals of the same image to self-supervise our feature representations. In contrast to approaches that define a global background consistency across images of a dataset, we use independent background consistency for each image, which allows for variation in backgrounds within the dataset. We leverage the region proposal network and minimize the differences between background representations of each predicted instance. See Figure~\ref{fig:systems}.

We accomplish this through the two step framework of Mask R-CNN. DARCNN utilizes self-supervision during training through first finding the top region proposals with confidence over threshold $k$ from the region proposal network. It passes each high confidence region proposal through the class agnostic mask head and determines which parts of the predicted instance are background. To do this, outputs from the mask head are passed through a sigmoid activation $\sigma$ and all values less than threshold $i$ are taken as background.

Then, we retrieve features output from our convolutional encoder $E_p^t$ that corresponds to background predictions. We minimize the differences between these background representations. Letting $r$ be a predicted region from shared region proposal network $R$ and and $M^t$ be our target mask prediction head, we define the background features $b(r)$ in Equation~\ref{eqn:target_loss}, where the indicator function extracts parts of $h_p^t$, domain specific features, that spatially corresponds to background mask predictions, threshold by value $i$ after sigmoid function $\sigma$. We can then define $\mu_b$, computed per image, as the mean of background representations across all regions in an image. These regions are predicted by region proposal network $R$ after taking domain invariant features $h_c^t$ as inputs. Finally, we define $L_{\textrm{target}}$ to minimize differences between background features. 

\vspace{-1em}
\begin{align}\label{eqn:target_loss} 
    b(r) &= \mathbbm{1}{[\sigma(M^t(r)) < i]} h_p^t \\
    \mu_b &= \frac{1}{N_{p}} \sum_{p \in R(h_c^t)} b(p)  \\
    L_{\textrm{target}} &= \frac{1}{N_{r}} \sum_{r \in R(h_c^t)}  \mid b(r) - \mu_b \mid
\end{align}




In our training process, the combination of the fully supervised and self-supervised losses from the source and target dataset respectively allows DARCNN to learn semantically relevant proposals and mask predictions.





\subsection{Augmented Pseudo-Labelling}
\label{section:pseudo_labelling}
Our first stage DARCNN utilizes feature-level adaptations for unsupervised domain adaptation and leads to initial coarse mask predictions that overcome large domain shift. However, it lacks strong image-level supervisory signal as in \cite{hoffman2018cycada} or \cite{chen2018domain}; the lack of a pixel-level signal leads to more unstable and unrefined segmentations. Therefore we propose a second stage image-level pseudo-labelling with our first stage DARCNN's output as pseudo-labels in order to gain this image-level supervision. See part b) in Figure~\ref{fig:systems}. Only the target branch of DARCNN is trained during the second stage pseudo-labelling process, while the source branch is frozen and no longer needed.

Canonical use of pseudo-labels depends on some amount of labelled data, however, as our method is unsupervised, we instead use high confidence predictions from our first stage DARCNN. We use confidence threshold $z$ to determine which labels to retrieve from the predictions. 

In addition, to better learn invariances of labels despite imaging conditions, including quality and noise, we apply data augmentation procedures to strengthen pseudo-labels from the first stage DARCNN. The augmentations ensure that the same instance segmentations will be predicted of a given input regardless of lighting, contrast, and blur.

The target branch of DARCNN is the final model to be used for unsupervised instance segmentation.

\section{Experiments}
In our experiments, we show that we are able to overcome limitations of past work to adapt between large domain shifts, as well generalize across many biomedical datasets. We quantitatively demonstrate DARCNN's performance on a large domain shift from COCO \cite{lin2014microsoft} to multiple biomedical datasets, bypassing drawbacks of previous literature that focuses on small domain shifts and specific datasets. In addition, in order to directly compare against existing work, we also show comparable performance to prior methods on tasks with small domain shifts following canonical biomedical adaptations of \cite{liu2020unsupervised}.

Our work is not limited to that of biomedical datasets, and is designed for all datasets with consistent backgrounds. We use biomedical datasets due to prevalence of homogeneous backgrounds in the biomedical field, for evaluation following previous unsupervised instance segmentation work \cite{liu2020unsupervised}, and to better illustrate our approach on large domain shifts from generalized COCO to diverse, biomedical datasets.

\subsection{Implementation details}
We evaluate our experiments on Aggregated Jaccard Index (AJI) \cite{kumar2017dataset}. AJI is used by prior work to evaluate the performance of instance segmentation; it computes an aggregated intersection cardinality numerator, and an aggregated union cardinality denominator for all ground truth and segmented predictions under consideration. It is a unified metric that measures both object-level and pixel-level performance, and is more stringent than other canonical metrics such as IOU. In addition, we also show pixel F1 score and object F1 score to measure performance in specific aspects. The annotations for all five target biomedical datasets are not used during unsupervised training of DARCNN, only for evaluation. For our experiments, we stop our model training $0.1$ epochs before loss plateaus.

We use Pytorch and build on the Detectron2 \cite{wu2019detectron2} framework, using the ResNet backbone. For our loss function $L_{\textrm{DARCNN}}$, we set $\alpha$ to increase over the first $0.1$ epochs to $1$, $\beta = 1$ and $\gamma = 0.1$. We use $k = 0.5$ as the confidence threshold for top predicted regions in our self-supervised loss, and $i = 0.5$ as our threshold for background. Confidence threshold $z$ is set to be $0.5$ for pseudo-labelling. Augmentation parameters used are Gaussian blur with sigma as $1$, and contrast and brightness are changed through scaling and delta factors $1.5$ and $-150$ respectively. We use learning rate $0.0001$, and vary maximum number of detections to return per image during inference through initial coarse inspection of training images. We choose the number to be $100$ or $50$ accordingly for each dataset. 

\subsection{Adaptation from Microscopy to Histopathology}
To compare against prior unsupervised domain adaptation methods tackling small domain shifts between biomedical images, we first quantitatively evaluate adaptation from a fluorescence microscopy dataset, BBBC \cite{ljosa2012annotated}, to two histopathology datasets, Kumar \cite{kumar2017dataset} and TNBC \cite{naylor2018segmentation}. This comparison follows that of Liu et al. \cite{liu2020unsupervised}, and we follow the same implementations of prior work and evaluation.

Importantly, we also demonstrate DARCNN's strong performance when adapting from a common dataset, COCO \cite{lin2014microsoft}, to the same two histopathology datasets. We demonstrate that even without a similar source biomedical dataset that may be difficult to obtain, we are still able to conduct unsupervised instance segmentation adapting from COCO.

We first preprocess our source dataset, BBBC. A total of $100$ training images and $50$ validation images from BBBC are used, following the official data split. $10,000$ patches of BBBC in size 256x256 are randomly cropped from the $100$ training images, and pixel values are inverted to better synthesize histopathology images following \cite{liu2020unsupervised}. Both Kumar \cite{kumar2017dataset} and TNBC \cite{naylor2018segmentation}, our target datasets, are trained with $10,000$ patches of size 256x256 without any labels, and evaluated on the specified test set. To compare our method against the nucleus specific methods of \cite{hou2019robust} and \cite{liu2020unsupervised}, we utilize the standalone, non-deep learning based unsupervised synthesis module of \cite{hou2019robust} as additional input. 

\begin{table}[htb]
\begin{center}
\begin{tabular}{|l|c|c|c|}
\hline
Method & AJI & Pixel-F1 & Object-F1 \\
\hline\hline
Chen et al. \cite{chen2018domain} & 0.4407 & 0.6405 & 0.6289 \\
DDMRL \cite{kim2019diversify} & 0.4642 & 0.7000 & 0.6872 \\
SIFA \cite{chen2019synergistic} & 0.4662 & 0.6994 & 0.6698 \\
CyCADA \cite{hoffman2018cycada} & 0.4721 & 0.7048 & 0.6866 \\
Hou et al. \cite{hou2019robust} & 0.4775 & 0.7029 & 0.6779 \\
Liu et al. \cite{liu2020unsupervised} & 0.5672 & 0.7593 & 0.7478 \\
\hline
\textbf{Ours from BBBC} & 0.5120 & 0.7175 & 0.6436 \\
\textbf{Ours from COCO} & 0.4906 & 0.6998 & 0.6396
 \\
\hline
\end{tabular}
\end{center}
\caption{Comparison of unsupervised methods adapting BBBC to TNBC. We see that DARCNN shows strong performance against prior work both with BBBC as source and with COCO as source, even against methods designed for nucleus-specific segmentation \cite{hou2019robust, liu2020unsupervised} and small domain shifts \cite{chen2018domain, kim2019diversify, chen2019synergistic, hoffman2018cycada}.}
\label{table:tnbc}
\vspace{-0.5em}
\end{table}

\begin{table}[htb]
\begin{center}
\begin{tabular}{|l|c|c|c|}
\hline
Method & AJI & Pixel-F1 & Object-F1 \\
\hline\hline    
Chen et al. \cite{chen2018domain} & 0.3756 & 0.6337 & 0.5737 \\
SIFA \cite{chen2019synergistic} & 0.3924 & 0.6880 & 0.6008 \\
CyCADA \cite{hoffman2018cycada} & 0.4447 & 0.7220 & 0.6567 \\
DDMRL \cite{kim2019diversify} & 0.4860 & 0.7109 & 0.6833 \\
Hou et al. \cite{hou2019robust} & 0.4980 & 0.7500 & 0.6890 \\
Liu et al. \cite{liu2020unsupervised} & 0.5610 & 0.7882 & 0.7483 \\
\hline
\textbf{Ours from BBBC} & 0.4461 & 0.6619 & 0.5410 \\
\textbf{Ours from COCO} & 0.4421 & 0.6549 & 0.5104 \\
\hline
\end{tabular}
\end{center} 
\caption{Comparison of unsupervised methods adapting BBBC to Kumar. DARCNN shows comparable performance to prior works on the nucleus segmentation task though it is designed to over-segment and retrieve all instances of interest.}
\label{table:kumar}
\vspace{-1em}
\end{table} 

We can see in Table~\ref{table:tnbc} that for the TNBC dataset in the scenario of a small domain shift, our method with fluorescence microscopy as source outperforms all prior work aside from \cite{liu2020unsupervised}. \cite{liu2020unsupervised} utilizes specific nuclei inpainting, which yields impressive performance, but can only be used for nuclei segmentation tasks.

More importantly, we show that DARCNN with COCO as the source dataset is able to achieve similar performance, also outperforming other methods aside from \cite{liu2020unsupervised}. Without labels from BBBC, we are still able to adapt from COCO to histopathology datasets, which is essential in cases where similar labelled biomedical datasets do not exist.

\begin{figure}[t]
\begin{center}
   \includegraphics[width=0.9\linewidth]{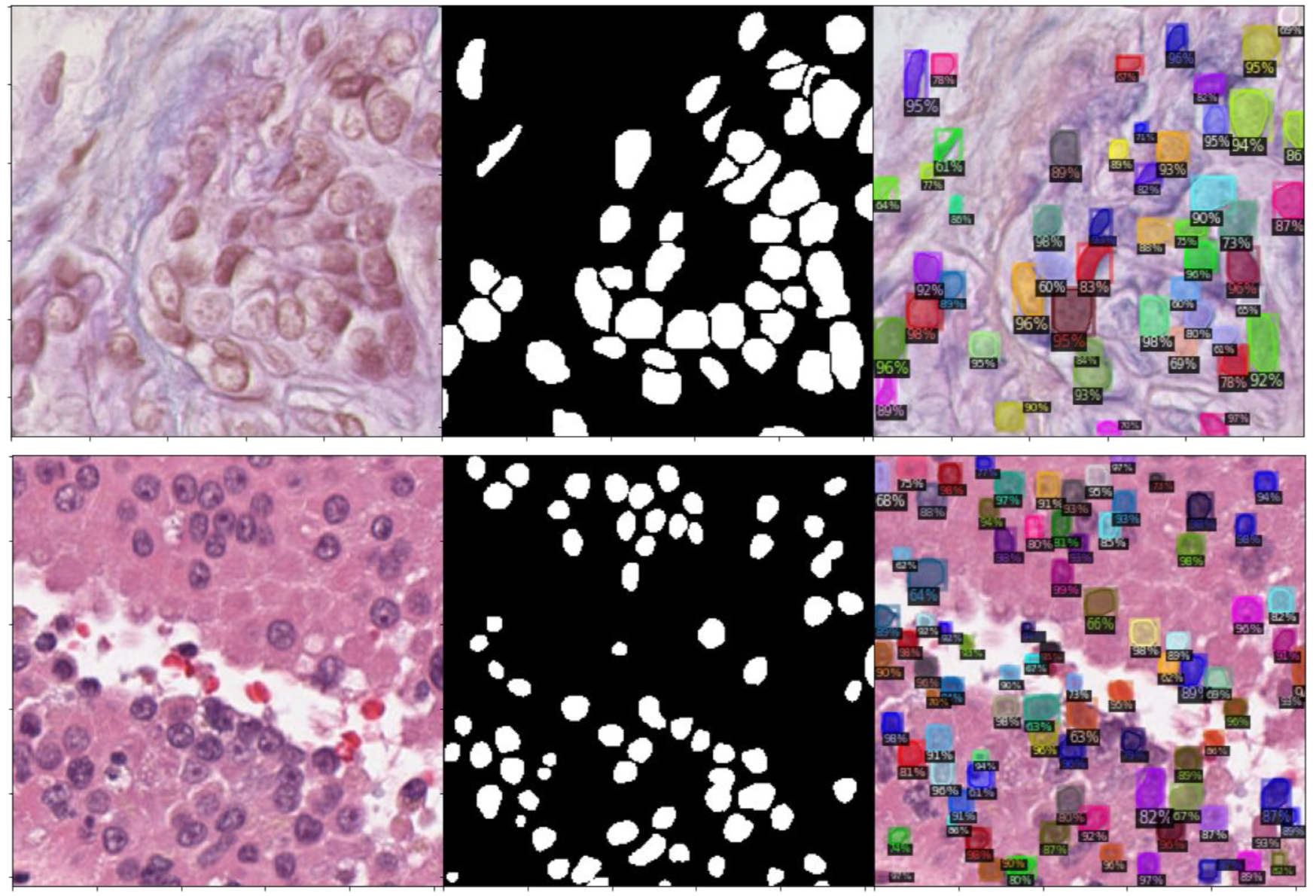}
\end{center}
\label{tnbc_fig}
   \caption{Qualitative results on DARCNN, adaptation from COCO to TNBC (top) \& Kumar (bottom) respectively. Left is input, middle is ground truth, and right shows instance segmentation result.}
\label{fig:kumar_results}
\vspace{-0.5em}
\end{figure}

In Table~\ref{table:kumar}, we see that for the Kumar dataset, due to DARCNN predicting more objects than are classified as nucleus, our scores are comparable but do not beat all of prior methods. We hypothesis that this is because past work focuses on image-level adaptation at small domain shifts, and their models are able to better learn what the object of focus is for segmentation. DARCNN is designed for instance segmentation of all distinct objects that exist within an image, hence predicts false positives from the nucleus perspective. See below Figure~\ref{fig:kumar_results} for an example of DARCNN adapted from COCO to Kumar; note that the red objects in the middle are not considered nucleus from the ground truth, but are segmented by DARCNN as objects of interest. DARCNN is useful for discovering objects in data when no labels are available, and produces class agnostic instance segmentations that can help scientists uncover new objects in complex data. Postprocessing methods could allow us to gain more insight into objects found by DARCNN. Through rule-based filtering with a coarse prior (filtering for masks with average pixel value less than a threshold, over a set of thresholds), our method obtains AJI = 0.5026, Pixel-F1 = 0.7272, Object-F1 = 0.6481, outperforming all prior work except \cite{liu2020unsupervised}, which is designed for nuclei segmentation.

\subsection{Ablation Studies}
We conduct ablation studies on DARCNN to showcase the effectiveness of the domain similarity loss, the background representation consistency loss, and the augmented pseudo-labelling stage. The DARCNN here is shown with COCO \cite{lin2014microsoft} as the source dataset, tackling our stated problem of large domain shift to biomedical datasets.

We perform this ablation study for two separate setups --- first we study DARCNN adapted to TNBC, trained with the unsupervised, standalone synthesis module of \cite{hou2019robust}, which allows for its comparable performance to previous nucleus-specific methods. Then we observe DARCNN's ablation performance for the BBBC dataset, showcasing each component's contribution to DARCNN without initial object-specific synthesis.


In Table~\ref{table:ablation_tnbc} and Table~\ref{table:ablation_bbbc}, we observe that class agnostic Mask R-CNN trained on COCO images performs extremely poorly on biomedical datasets. We can also see that for the first stage of DARCNN, both the domain similarity loss and representation consistency loss improved model performance. Especially in the case of BBBC without the unsupervised synthesis module, our representation consistency loss dramatically improves the performance of DARCNN due to ability to gain self-supervised signal for features. The full first stage DARCNN also shows significantly higher performance in both datasets than the initial baselines.

\begin{table}[htb]
\begin{center}
\small
\begin{tabular}{|l|c|c|c|}
\hline
Method & AJI & Pixel-F1 & Object-F1 \\
\hline\hline
\multicolumn{4}{|c|}{Mask R-CNN} \\
\hline
w/ COCO pre-trained & 0.0060 & 0.2769 & 0.0181 \\
w/ synthesized images & 0.3332 & 0.5782 & 0.6061 \\
\hline
\multicolumn{4}{|c|}{First stage DARCNN} \\
\hline
Domain sim. only & 0.3687 & 0.6023 & 0.6099 \\
Bg. consistency only & 0.3808 & 0.6120 & 0.5470 \\
Full 1st stage DARCNN & 0.4071 & 0.6353 & 0.5986 \\
\hline
\multicolumn{4}{|c|}{Second stage DARCNN} \\
\hline
Pseudo-label w/o aug & 0.4463 & 0.6781 & 0.6339 \\
\textbf{Full 2nd stage DARCNN} & 0.4906 & 0.6998 & 0.6396 \\
\hline
\end{tabular}
\end{center}
\caption{Ablation study adapting from COCO as source to TNBC.}
\label{table:ablation_tnbc}
\vspace{-0.5em}
\end{table}

\begin{table}[htb]
\begin{center}
\small
\begin{tabular}{|l|c|c|c|}
\hline
Method & AJI & Pixel-F1 & Object-F1 \\
\hline\hline
\multicolumn{4}{|c|}{Mask R-CNN} \\
\hline
w/ COCO pre-trained & 0.0315 & 0.3144 & 0.0818 \\
\hline
\multicolumn{4}{|c|}{First stage DARCNN} \\
\hline
Domain sim. only & 0.1414 & 0.4905 & 0.4295 \\
Bg. consistency only & 0.3250 & 0.7128 & 0.5720 \\
Full 1st stage DARCNN & 0.3371 & 0.6409 & 0.5904 \\ 
\hline
\multicolumn{4}{|c|}{Second stage DARCNN} \\
\hline
Pseudo-label w/o aug & 0.4349 & 0.6914 & 0.7151 \\
\textbf{Full 2nd stage DARCNN} & 0.4725 & 0.6586 & 0.6733 \\
\hline
\end{tabular}
\end{center}
\caption{Ablation study adapting from COCO as source to BBBC.}
\label{table:ablation_bbbc}
\vspace{-1em}
\end{table}

In the second stage of DARCNN, pseudo-labelling also improved our unsupervised instance segmentation performance. Similarly, the performance improvement from pseudo-labelling in the BBBC case without the synthesis module is larger than in TNBC, as for BBBC it is the first time DARCNN's target branch receives image-level supervision. Pseudo-labelling with augmented data demonstrates even better performance, helping remove difficulties in segmentation under various imaging conditions of biomedical datasets. We also show in Figure~\ref{fig:kumar_results} a qualitative example of DARCCN adapting from COCO to TNBC, capturing nuclei as objects of interests in the image.


\subsection{Adaptation from COCO to Additional Biomedical Datasets}
Most importantly, we demonstrate DARCNN's ability to generalize across datasets by comparing performance on COCO adapted to three diverse biomedical datasets -- the fluorescence microscopy dataset \cite{ljosa2012annotated}, cryogenic electron tomography dataset \cite{gubins2020shrec}, and brain MRI dataset \cite{menze2014multimodal, bakas2017advancing, bakas2018identifying}. Through this, we demonstrate potential for object discovery without the need for similar labelled datasets to the target domain, and without need for designing object-specific models for segmentation of particular biomedical datasets. 

We compare our results with the prior methods that can be used outside of specific biomedical datasets. \cite{chen2019synergistic, hou2019robust, liu2020unsupervised} depend on specific biomedical images and nuclei synthesis methods, hence we do not use them as comparison.

\begin{table}[htb]
\begin{center}
\small
\begin{tabular}{|l|c|c|c|c|}
\hline
Dataset & Method & AJI & Pixel-F1 & Object-F1 \\
\hline\hline
BBBC & Chen et al. \cite{chen2018domain} & 0.1500 & 0.5251 & 0.2111 \\
& CyCADA \cite{hoffman2018cycada} & 0.1231 & 0.5173 & 0.1917 \\
& DDMRL \cite{kim2019diversify} & 0.0928 & 0.5188 & 0.1145 \\
& \textbf{Ours} & 0.4725 & 0.6586 & 0.6733 \\
\hline
SHREC & Chen et al. \cite{chen2018domain} & 0.0 & 0.0 & 0.0 \\
& CyCADA \cite{hoffman2018cycada} & 0.0051 & 0.0064 & 0.0025 \\
& DDMRL \cite{kim2019diversify} & 0.0039 & 0.0182 & 0.0 \\
& \textbf{Ours} & 0.1268 & 0.3007 & 0.3371 \\
\hline
BraTS & Chen et al. \cite{chen2018domain} & 0.2868* & - & - \\
& CyCADA \cite{hoffman2018cycada} & 0.3485* & - & - \\
& DDMRL \cite{kim2019diversify} & 0.3951* & - & - \\
& \textbf{Ours} & 0.5577* & - & - \\
\hline
\end{tabular}
\end{center} 
\caption{Comparison of unsupervised methods adapting from COCO to fluorescence microscopy (BBBC), cryogenic electron tomography (SHREC), and radiology (BraTS) datasets. *Indicates metric measuring maximum intersection over union.}
\vspace{-1em}
\end{table}

DARCNN significantly outperforms all other methods when adapting with a large domain shift from COCO. As \cite{hoffman2018cycada, kim2019diversify} depend on CycleGAN \cite{zhu2017unpaired}, and \cite{chen2018domain} depends on image-level shift, we hypothesize that these methods perform poorly when tasked with a large domain shift from COCO to natural biomedical images. DARCNN is able to generalize across diverse biomedical datasets and adapt between common objects to objects in microscopy, tomography, and MRI due to its two stage sequential feature-level adaptation and image-level pseudo-labelling.

\begin{figure}[t]
\begin{center}
   \includegraphics[width=0.9\linewidth]{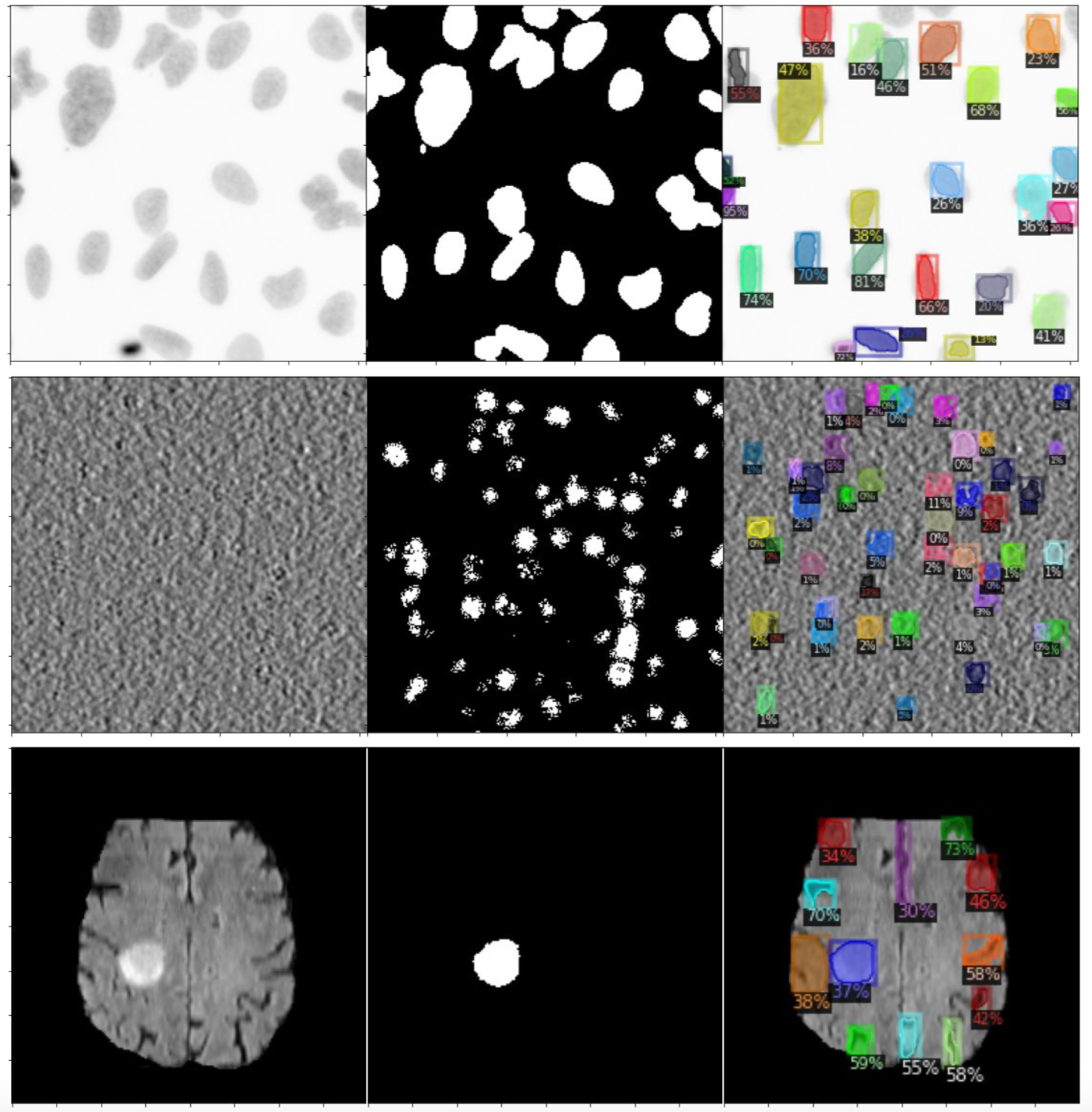}
\end{center}
\vspace{-0.5em}
\label{tnbc_fig}
   \caption{Qualitative results on DARCNN, adaptation from COCO to BBBC (top), SHREC (middle), and BraTS (bottom) datasets.}
\label{fig:additional}
\vspace{-1.5em}
\end{figure}

For BBBC, DARCNN is able to significantly outperform prior work as the background consistency assumption is strong. In addition, even for a more challenging task like instance segmentation in SHREC where the signal to noise ratio is low, DARCNN is still able to capture objects of interests given noisy background. Prior work, CyCADA \cite{hoffman2018cycada}, Chen et al. \cite{chen2018domain}, and DDMRL \cite{kim2019diversify}, are all unable to learn meaningful adaptations from COCO to SHREC due to the difficulties of image-level adaptation when even human recognition is limited. In the BraTS dataset, due to the specificity of the detected object, tumor, we measure performance by maximum intersection over union from closest predicted object. Though this does not account for unbounded false positives, we provide a qualitative example of performance to supplement. In Figure~\ref{fig:additional}, we see that DARCNN also picks up on ridges and darker spots in the BraTS MRI, which could potentially be useful in understanding additional structures of interest. We show qualitative examples on our three biomedical datasets in Figure~\ref{fig:additional}.

\subsection{Adaptation from COCO to CryoET}
Finally, we showcase the promise of our unsupervised instance segmentation model for adapting from COCO to an unlabelled cryogenic electron tomography (cryoET) dataset, collected by Dr. Wah Chiu's group at SLAC National Accelerator Laboratory. This cryoET dataset contains tomograms of crowded cellular environments in which objects are too dense and of too underexplored a subject area to be annotated. We qualitatively evaluate the performance of DARCNN instance segmentations in this dataset.

\begin{figure}[htb]
\vspace{-0.5em}
\begin{center}
   \includegraphics[width=0.8\linewidth]{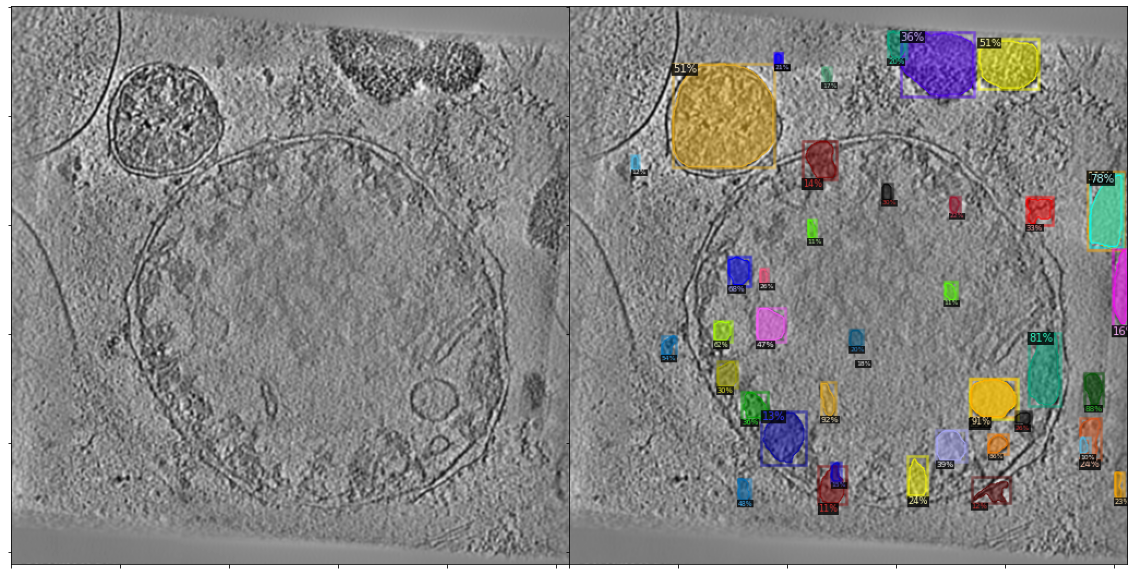}
\end{center}
   \caption{Qualitative results on DARCNN, adaptation from COCO to cryoET, demonstrating discovery of distinct objects of interests.}
\label{fig:cryoem}
\vspace{-1em}
\end{figure}

In Figure~\ref{fig:cryoem}, our unsupervised algorithm segments known biological objects such as the autophagosome and granules. In addition, DARCNN also discovers distinct vesicles and organelles inside the amphisome, difficult for humans to annotate, and representing potential new objects of interest. 


\vspace{-0.5em}
\section{Conclusion}
We propose DARCNN, a two stage feature-level adaptation and image-level pseudo-labelling method for unsupervised instance segmentation. We leverage the abundance of labelled benchmark datasets for domain adaptation to unlabelled biomedical images. DARCNN tackles large domain shifts between common and biomedical objects, and can be used across diverse datasets with consistent background. Through a domain separation module, a representation consistency loss, and augmented pseudo-labelling, we achieve strong performance in multiple experiments as well as show potential for object discovery within biomedical datasets.

\textbf{Acknowledgments.} We thank support from the Chan-Zuckerberg Initiative (to J.H., S.Y. and W.C.) and National Institutes of Health (grant number P01NS092525 to W.C.).

{\small
\bibliographystyle{ieee_fullname}
\bibliography{egbib}
}

\end{document}